\documentclass[11pt,a4paper]{article}
\usepackage[hyperref]{eacl2021}
\usepackage{times}
\usepackage{latexsym}

\usepackage{microtype}

\aclfinalcopy 

\usepackage{amsmath,amsfonts,bm}

\def\eqref#1{equation~\ref{#1}}

\def\1{\bm{1}}

\DeclareMathAlphabet{\mathsfit}{\encodingdefault}{\sfdefault}{m}{sl}
\SetMathAlphabet{\mathsfit}{bold}{\encodingdefault}{\sfdefault}{bx}{n}

\newcommand{\Ls}{\mathcal{L}}

\usepackage{mathabx}
\usepackage{subfig}
\usepackage{float}
\usepackage{times}
\usepackage{soul}
\usepackage{url}
\usepackage{hyperref}
\usepackage[utf8]{inputenc}
\usepackage{caption}
\usepackage{graphicx}
\usepackage{multirow}
\usepackage{amsmath}
\usepackage{amsthm}
\usepackage{booktabs}
\usepackage{algorithm}
\usepackage{algorithmic}
\usepackage[normalem]{ulem}
\useunder{\uline}{\ul}{}

\title{Joint Energy-based Model Training for Better Calibrated \\ Natural Language Understanding Models}

\author{Tianxing He \\
  MIT \\
  \texttt{cloudygoose@csail.mit.edu} \\\And
  Bryan McCann\thanks{\ \ \ Bryan McCann contributed to this work while he was at Salesforce Research.} \\
  Salesforce Research \\
  \texttt{bryan@you.com} \\
  \AND
  Caiming Xiong \\
  Salesforce Research \\
  \texttt{cxiong@salesforce.com} \And
    Ehsan Hosseini-Asl\\
  Salesforce Research\\ 
  \texttt{ehosseiniasl@salesforce.com} 
}
  
\date{}

\begin{document}
\maketitle
\begin{abstract}
In this work, we explore joint energy-based model (EBM) training during the finetuning of pretrained text encoders (e.g., Roberta) for natural language understanding (NLU) tasks. Our experiments show that EBM training can help the model reach a better calibration that is competitive to strong baselines, with little or no loss in accuracy. We discuss three variants of energy functions (namely \textit{scalar}, \textit{hidden}, and \textit{sharp-hidden}) that can be defined on top of a text encoder, and  compare them in experiments. Due to the discreteness of text data, we adopt noise contrastive estimation (NCE) to train the energy-based model. To make NCE training more effective, we train an auto-regressive noise model with the masked language model (MLM) objective. We release our code at \url{https://github.com/salesforce/ebm_calibration_nlu}.
\end{abstract}

\section{Introduction}

Calibration refers to how well a classification model's confidence (reflected by its output posterior probability) aligns with its actual accuracy. As deep learning models achieve amazing accuracy in computer vision \citep{He2015resnet} or natural language processing (NLP) \citep{yinhan19roberta, jacob18bert}, more research attention has been drawn to the calibration aspect of these models. As shown by \citet{pmlr-v70-guo17a}, the high accuracy from deep models does not always lead to better calibration. This motivates an important line of works attempting to achieve a better trade-off between accuracy and calibration.


Most existing calibration methods \citep{pmlr-v70-guo17a,kuma19verifiedcal,Zadrozny01obtainingcalibrated} generally rescale the posterior distribution predicted from the classifier after training. Such post-processing methods require a held-out development set with a decent number of samples to be available. To overcome this constraint, \citet{jung-etal-2020-posterior} uses a penalty term to encourage better calibration during training. 

In another line of work, \citet{grathwohl2019classifier} shows that one can jointly train an energy-based model (EBM) during the standard training of a neural classifier. Although calibration is not explicitly addressed during EBM training, the calibration of the resulting model is shown to be greatly improved. Some intuitions of the underlying reasons will be given in Section \ref{sec:ncetraining}. However, the training framework proposed by \citet{grathwohl2019classifier} is designed for image classifiers, and it can not be readily applied to discrete text data.

In this work, we propose a framework that uses noise contrastive estimation (NCE) to jointly train an energy-based model during the finetuning of pretrained text encoders (e.g., BERT \citep{jacob18bert} or Roberta \citep{yinhan19roberta}) for NLU tasks. We compare several variants of energy functions that can be defined on top of the encoder. Our experiments show that the resulting models achieve competitive calibration results comparing to strong baselines, with little or no loss in accuracy.

\section{Framework}

\subsection{Notations and Background}
\label{sec:notations}

We focus on the finetuning of pretrained text encoder on NLU tasks. We assume samples from the data distribution $P_D$ are in the form of $(x,y)$ pairs, where $x$ usually refers to a single or a pair of sentences, and $y$ refers to the corresponding label. The number of classes are denoted by $|Y|$.

Given input $x$, we first use a text encoder model (e.g., BERT or Roberta) to encode it and we denote this embedding as $\text{enc}(x)$. For the target classification task, a classifier $f_{\text{CLS}}$, which could be a simple linear transform or a multi-layer perception (MLP), will be applied to $\text{enc}(x)$. We denote the output logits as $f_{\text{CLS}}(\text{enc}(x))$, whose dimension is equal to the number of possible classes $|Y|$. The $y$-th logit is denoted by $f_{\text{CLS}}(\text{enc}(x))[y]$. The posterior distribution $P_{\theta}(y|x)$ is obtained by applying a softmax operation to the logits, where $\theta$ refers to the parameters in the model.

In standard finetuning, the cross-entropy (CE) loss and gradient based optimizers are used to train the classifier:
\begin{equation}
    \mathcal{L}_\text{CE} = \mathop{\mathbb{E}}_{(x, y) \sim P_{D}} (-\log P_{\theta}(y|x)).
\end{equation}
In the next few sections, we discuss how we define and jointly train an EBM on top of the text encoder.

\subsection{Definitions of Energy Function}
\label{sec:energy_func_def}

An energy-based model \citep{LeCun06atutorial} expresses $P_{\theta}(x)$ as:
\begin{equation}
    P_{\theta}(x) = \frac{\exp(-E_{\theta}(x))}{Z},
\end{equation}
where $Z$ is the normalization factor, and is usually intractable to compute. We refer to $E_{\theta}(x)$, which returns a scalar value, as \textit{the energy function}. We now define three variants of energy functions.

\textbf{Variant \textit{scalar}:} We introduce another linear layer $g_{\text{S}}$ whose output is a scalar. And we use it to define the energy function:
\begin{equation}
    \hat{E}_{\theta}(x)=g_{\text{S}}(\text{enc}(x)).
\end{equation}

\begin{figure}
    \centering
    \includegraphics[width=\linewidth]{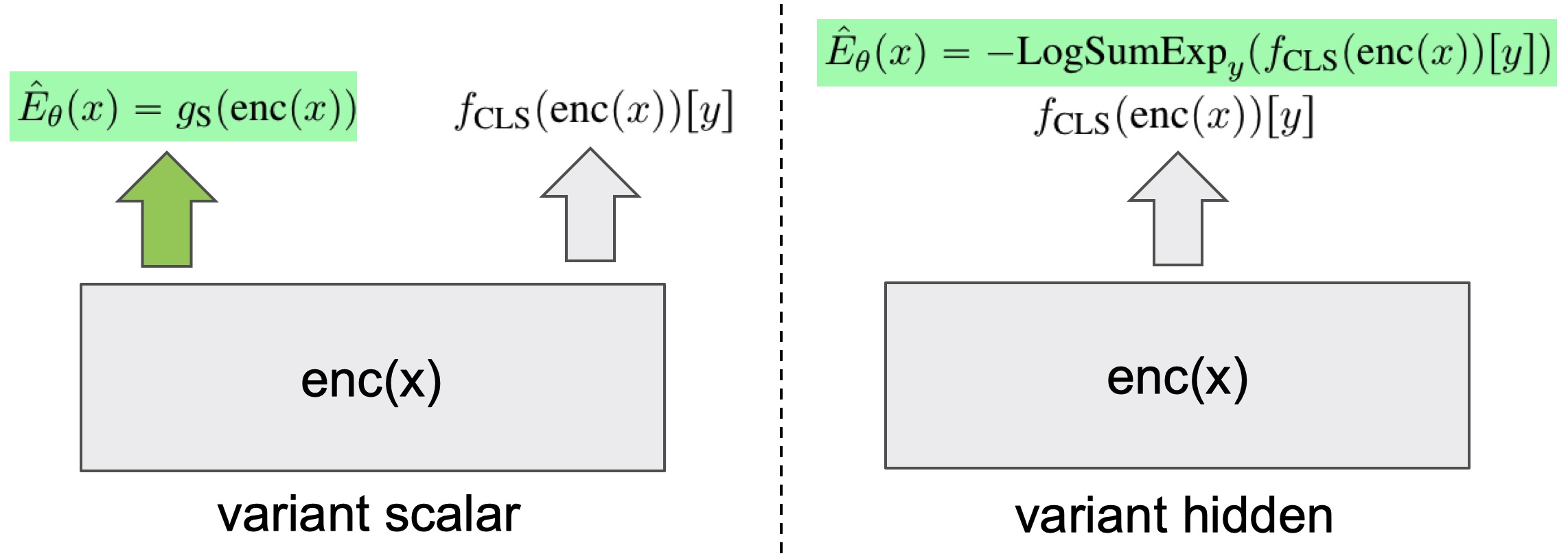}
        \vspace{-0.2cm}
    \caption{Comparison of the \textit{scalar} and the \textit{hidden} variants of energy functions. The modules introduced for EBM are shaded in green.}
    \label{fig:energy_version_compare}
\end{figure}

\textbf{Variant \textit{hidden}:} As pointed out by \citet{grathwohl2019classifier}, there's an EBM ``hidden'' in every neural classifier with softmax output, and the energy function for $x$ can be derived\footnote{Please see Appendix \ref{appsec:hidden_derivation} for the detailed derivation.} as:
\begin{equation}
\label{eq:hidden}
    \hat{E}_{\theta}(x)=-\text{LogSumExp}^{|Y|}_{y=1}(f_{\text{CLS}}(\text{enc}(x))[y]).
\end{equation}
Different from the \textit{scalar} variant, here the energy function directly uses the logits for prediction (visualized in Figure \ref{fig:energy_version_compare}). Hence the impact on the model's classification behavior could be larger.

\textbf{Variant \textit{sharp-hidden}:} 
The \textit{hidden} variant has a potential weakness that, the correlation between input $x$ and the prediction $y$ is not addressed because the energy is distributed among all the logits.  Motivated by this potential issue, we propose the following ``sharp'' variant:
\begin{equation}
\label{eq:sharp-hidden}
    \hat{E}_{\theta}(x)=-\max_y f_{\text{CLS}}(\text{enc}(x))[y].
\end{equation}
Note that (\ref{eq:sharp-hidden}) can be viewed as an approximation to (\ref{eq:hidden}), and we find it to work well in practice.

Finally, for each variant, we define the energy function to be $E_{\theta}(x)=\hat{E}_{\theta}(x) - \log P_N(x)$, where $P_N$ is the \textit{noise distribution} introduced for NCE. We will motivate this design choice below.


\subsection{NCE Training}
\label{sec:ncetraining}
We use noise contrastive estimation (NCE) \citep{pmlr-v9-gutmann10a,ma-collins-2018-noise} to jointly train the energy model. NCE trains the model to discriminate between data samples and noise samples from a given noise distribution $P_N$. We formulate the NCE loss below:
\begin{equation}
\small
\begin{split}
    \Ls_\text{NCE} = \mathop{\mathbb{E}}_{x_+ \sim p_D} - \log \frac{\tilde{P}_{\theta}(x_+)}{\tilde{P}_{\theta}(x_+)+K\cdot P_N(x_+)} + \\ K \cdot \mathop{\mathbb{E}}_{x_- \sim p_N} - \log \frac{K \cdot P_N(x_-)}{\tilde{P}_{\theta}(x_-)+K\cdot P_N(x_-)},
\end{split}
\end{equation}
where $K$ is the ratio of noise samples. Note that $\tilde{P}_{\theta}(x)$ does not need to be normalized by construction, therefore we set it to be $\tilde{P}_{\theta}(x)=\exp(-E_\theta(x))$. In our experiments, we mostly report results with noise ratio $K=8$, while in some cases we find that a small ratio of $K=1$ works slightly better. We have also tried with larger ratio such as 16, but the gain is minimal.

If we directly use the formulations of $\hat{E}_\theta(x)$ defined in last section as the energy function, the optimization will be difficult because of the $P_N(x)$ terms (which could be of very small value) in the NCE objective. To overcome this issue, we follow \citet{Deng2020Residual} and define $E_{\theta}(x)=\hat{E}_{\theta}(x) - \log P_N(x)$. In this way, the $P_N(x)$ terms are canceled, and the objective is simplified to:
\begin{equation}
\small
\begin{split}
    \Ls_\text{NCE} = 
    \mathop{\mathbb{E}}_{x_+ \sim p_D} - \log \frac{1}{1 + K \cdot \exp(\hat{E}_{\theta}(x_+))} + \\ K \cdot \mathop{\mathbb{E}}_{x_- \sim p_N} - \log \frac{K}{K + \exp(-\hat{E}_{\theta}(x_-))}.
    \end{split}
\end{equation}

In training, we jointly optimize $\Ls_\text{CE}$ and $\Ls_\text{NCE}$ with the Adam optimizer \citep{adam14kingma}:
\begin{equation}
    \Ls_\text{joint} = \Ls_\text{CE} + \Ls_\text{NCE}.
\end{equation}

Intuitively, joint EBM training makes the model aware of $P(x)$, instead of only focusing on predicting $P(y|x)$ as in standard finetuning. This awareness can potentially help with calibration because the model can be more conservative when it detects the input is out-of-distribution.


\subsection{Construction of Noise Distribution}
For the choice of noise distribution $P_N$, in our preliminary trials, we finetune the GPT-2 language model \citep{radford2019language} with samples from the target training set using the standard LM objective. However during NCE training, we find that the energy model can easily discriminate between data samples and noise samples, which makes training ineffective. To alleviate this issue, we adopt an objective similar\footnote{The difference is that we still train the model to generate the full sentence, instead of only the masked words.} to the masked language model (MLM) loss \citep{jacob18bert} during the finetuning of the noise model (GPT-2): With a given mask ratio $M$, we randomly mask part of $x$, and train the model to complete it:
\begin{equation}
\small
    \Ls_\text{MLM} = \mathop{\mathbb{E}}_{x \sim P_D, x^m \sim P_\text{mask}(x^m|x;M)} - \log P_\text{N} (x|x^m).
\end{equation}
During noise sample generation, adopting the same mask ratio $M$, we feed a masked $x^m$ to the LM ($x$ is from the training set), and use the generated sample as the noise sample. In this way, the noise distribution is made closer to the data distribution. In our experiments we set $M=0.4$. During generation, we use top-$k$ \citep{fan2018-storyhierarchical} sampling with $k=20$. More details are provided in Appendix \ref{appsec:lmnoise}.

\section{Experiments}

\paragraph{Setting}
 We consider finetuning the Roberta-base model\footnote{Our code is based on \url{https://github.com/huggingface/transformers}.}, on eight GLUE tasks \citep{wang-etal-2018-glue}. We do not include results on STS-B because it is a regression task. 
To measure calibration error, we follow \citet{jung-etal-2020-posterior, grathwohl2019classifier} and use the expected calibration error (ECE) metric with $B$ (number of bins) set to 20. To save space, we defer detailed definition of ECE to Appendix \ref{appsec:ece_def}.

For baseline or NCE training, we follow the recommended hyper-parameters (learning rate, batch size, etc.) for Roberta \citep{yinhan19roberta}. Since NCE training requires more computation (because of the noise ratio), we have tried finetuning the baseline with more steps, but we find that gives worse ECE and very little or no improvement on accuracy. 

We compare EBM training with three strong baselines for calibration: \textit{posterior calibrated training} (PosCal) \citep{jung-etal-2020-posterior}, \textit{temperature scaling} (T-Scal) \citep{pmlr-v70-guo17a},  and \textit{scaling-binning calibrator} (Scal-bin) \citep{kuma19verifiedcal}. For PosCal and Scal-bin, we use the published code.
 
Scal-bin and T-Scal require a development set for parameter learning and a test set for evaluation, but for each GLUE task we only have one labeled development set available. Therefore, in this work we treat half of the standard development set as test set, and keep the other half as development set.
 
 \begin{figure}
    \centering
    \includegraphics[width=\linewidth]{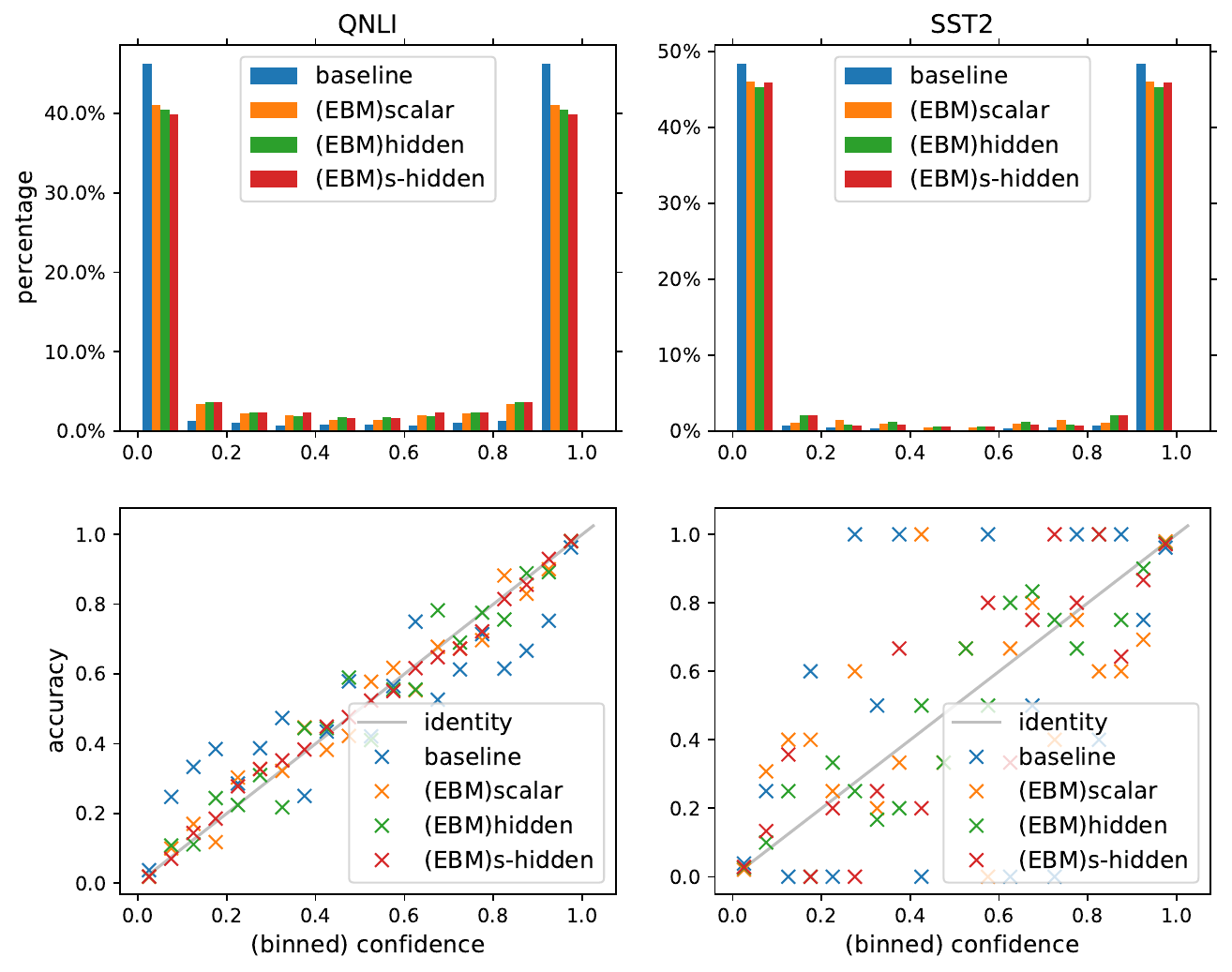}
        \vspace{-0.2cm}
    \caption{Visualization of calibration on QNLI and SST-2. In the histogram plots, we use 10 bins instead of 20 for better readability. An enlarged version of this figure is provided in Appendix \ref{appsec:auxiliary_results}.}
    \label{fig:cal_qnli_rte}
\end{figure}

\paragraph{Results}

\begin{table*}[t]
\centering
\small
\addtolength{\tabcolsep}{-3pt}
\begin{tabular}{|c|cc|cc|cc|cc|cc|cc|cc|cc|}
\hline
                           & \multicolumn{2}{c|}{\textbf{SST-2}} & \multicolumn{2}{c|}{\textbf{MNLI}} & \multicolumn{2}{c|}{\textbf{MNLI(mm)}} & \multicolumn{2}{c|}{\textbf{QNLI}} & \multicolumn{2}{c|}{\textbf{QQP}} & \multicolumn{2}{c|}{\textbf{MRPC}} & \multicolumn{2}{c|}{\textbf{CoLA}} & \multicolumn{2}{c|}{\textbf{Average}} \\
\textbf{Method}            & \textbf{acc.}    & \textbf{ECE}     & \textbf{acc.}    & \textbf{ECE}    & \textbf{acc.}      & \textbf{ECE}      & \textbf{acc.}    & \textbf{ECE}    & \textbf{acc.}   & \textbf{ECE}    & \textbf{acc.}    & \textbf{ECE}    & \textbf{mcc.}    & \textbf{ECE}    & \textbf{perf.}   & \textbf{ECE}   \\ \hline
\textbf{Baseline}          & .942             & .050             & .876             & .067            & .872               & .068              & .929             & .043            & .904            & .034            & .862             & .133            & .539             & .182            & .802             & .102           \\
\textbf{Scal-bin(train)}   & .940             & .036             & .872             & .051            & .869               & .056              & .931             & .034            & .904            & .035            & .843             & .092            & .586             & .146            & .791             & .096           \\
\textbf{T-Scal(train)}     & .942             & .042             & .876             & .058            & .872               & .060              & .929             & .030            & .904            & .034            & .862             & .126            & .539             & .175            & .802             & .096           \\
\textbf{PosCal}            & .944             & .040             & .876             & .067            & .872               & .067              & .930             & .039            & .905            & .032            & .867             & .129            & .540             & .184            & .810             & .092           \\
\textbf{(EBM)scalar}       & .942             & .033             & .871             & .038            & .871               & .047              & .927             & .016            & .899            & .034            & .862             & .098            & .540             & .150            & .801             & .073           \\
\textbf{(EBM)hidden}       & .956             & \textbf{.032}    & .869             & .032            & .868               & .044              & .923             & .016            & .900            & .033            & .867             & .099            & .545             & \textbf{.131}   & .807             & .063  \\
\textbf{(EBM)s-hidden} & .947             & .038             & .875             & \textbf{.027}   & .872               & \textbf{.031}     & .930             & \textbf{.016}   & .900            & \textbf{.032}   & .862             & \textbf{.089}   & .563             & .133            & .815             & .069           \\ \hline
\textbf{Scal-bin(dev)}     & .944             & .019             & .876             & .030            & .870               & .032              & .931             & .021            & .905            & .021            & .862             & .062            & .557             & .048            & .802             & .052           \\
\textbf{T-Scal(dev)}       & .942             & .037             & .876             & .024            & .872               & .026              & .929             & .018            & .904            & .026            & .862             & .126            & .539             & .109            & .802             & .072           \\ \hline
\end{tabular}
    \vspace{-0.2cm}
\caption{Test-set accuracy and ECE results for different methods on GLUE tasks. ``s-hidden'' refers to the \textit{sharp-hidden} variant. The leading zeros are omitted to save space. Note that T-Scal and Scal-bin are applied to the training set or the development set, respectively. Due to space constraint, results on RTE and WNLI are deferred to Table \ref{tab:res_rte_wnli}. The average value is compute on all nine test sets.  For each task, the method that achieves best calibration without using the development set are shown in bold.}
\label{tab:main_result}
\end{table*}

\begin{table}
\small
\centering
\addtolength{\tabcolsep}{-3pt}
\begin{tabular}{|c|cc|cc|}
\hline
\multicolumn{1}{|l|}{}   & \multicolumn{2}{c|}{\textbf{RTE}} & \multicolumn{2}{c|}{\textbf{WNLI}} \\
\textbf{Method}        & \textbf{acc.}   & \textbf{ECE}    & \textbf{acc.}    & \textbf{ECE}    \\ \hline
\textbf{Baseline}        & .724            & .279            & .571             & .058            \\
\textbf{Scal-bin(train)} & .717            & .271            & .457             & .144            \\
\textbf{T-Scal(train)}   & .724            & .275            & .571             & .063            \\
\textbf{PosCal}          & .789            & .206            & .571             & .060            \\
\textbf{(EBM)scalar}     & .753            & .207            & .542             & \textbf{.033}   \\
\textbf{(EBM)hidden}     & .797            & \textbf{.148}   & .542             & .036            \\
\textbf{(EBM)s-hidden}   & .811            & .182            & .571             & .073            \\ \hline
\textbf{Scal-bin(dev)}   & .731            & .042            & .542             & .189            \\
\textbf{T-Scal(dev)}     & .724            & .235            & .571             & .046            \\ \hline
\end{tabular}
\caption{(Following Table \ref{tab:main_result}) Main results on RTE and WNLI.}
\label{tab:res_rte_wnli}
\end{table}

In Table \ref{tab:main_result} and Table \ref{tab:res_rte_wnli} we compare test-set accuracy\footnote{For CoLA we report with Matthews correlation coefficient (mcc).} and ECE for different methods on the GLUE tasks. For fair comparison between Scal-bin / T-Scal and EBM training (which does not use the development set), we apply them to the whole training set. We also report their performance when applied to the development set for reference.

In most tasks, all three EBM variants get substantial improvement in ECE with little or no loss in accuracy comparing to the (strong) baseline methods. Moreover, the performance of EBM training is comparable to Scal-bin / T-Scal applied to the development set, while their performance degrades when the development set is not available. Among the three variants, on average, the \textit{sharp-hidden} variant achieves the best accuracy, while the \textit{hidden} variant achieves best calibration. We visualize the calibration error in Figure \ref{fig:cal_qnli_rte}. 

\begin{figure}
    \centering
    \includegraphics[width=\linewidth]{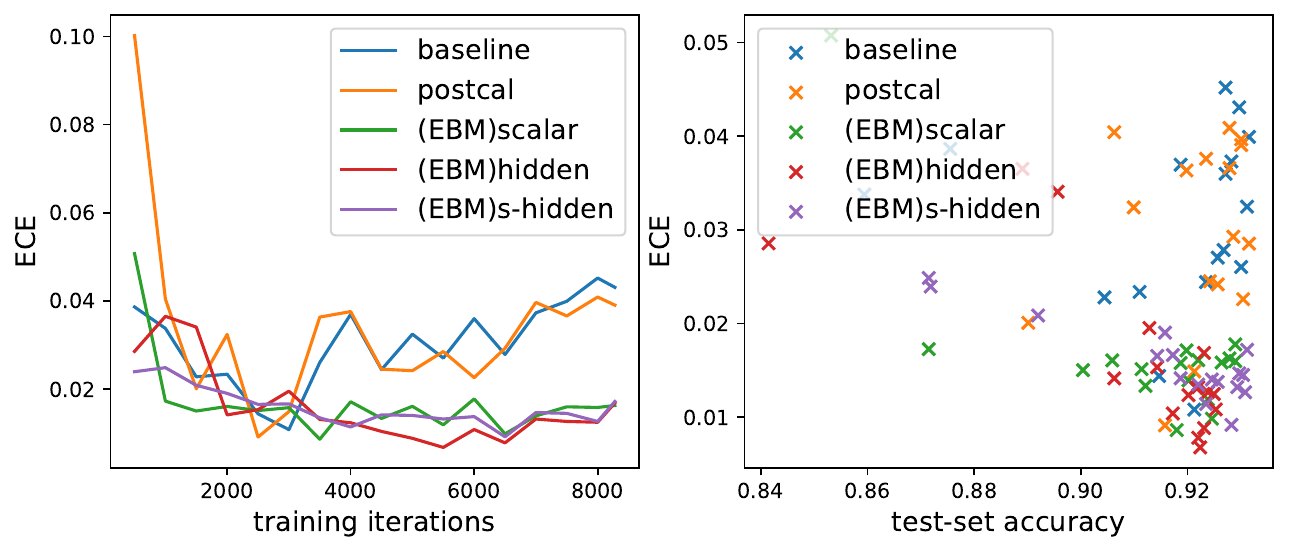}
    \vspace{-0.2cm}
    \caption{(QNLI) Left: How ECE changes during training. Right: The trade-off between accuracy and ECE for checkpoints (every 500 iterations) during training.} 
    \label{fig:traj_qnli}
\end{figure}

In Figure \ref{fig:traj_qnli}, we plot how test-set ECE changes during training. It is shown as the training reaches the high-accuracy area, the calibration for baseline model becomes worse, while EBM training is able to reach a better trade-off between accuracy and calibration.

\begin{figure}
    \centering
    \includegraphics[width=0.8\linewidth]{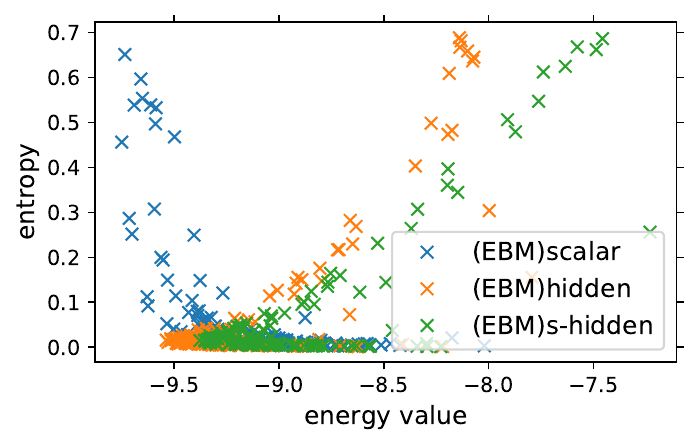}
    \vspace{-0.2cm}
    \caption{The entropy of the posterior ($P_\theta(\cdot|x)$) versus energy value $\hat{E}_\theta(x)$ for SST-2 test-set samples.}
    \label{fig:ent_eng_sst}
\end{figure}

How does the model get better calibration? In Figure \ref{fig:ent_eng_sst}, we compute and plot the energy value $\hat{E}_\theta(x)$ versus the entropy of the posterior distribution $\mathcal{H}(P_\theta(\cdot | x ))=\sum^{|Y|}_{y=1} -P_\theta(y|x)\log P_\theta(y|x)$, for samples in the SST-2 test set. It is shown that models trained with the \textit{hidden} and \textit{sharp-hidden} variants tend to assign more conservative predictions (reflected by higher entropy) for higher-energy (less likely) samples. We suspect this is due to the strong coupling between the energy function and the classification logits. We provide concrete examples in Table \ref{tab:energy_examples}. However, we need to mention that we do not observe this interesting trend (Figure \ref{fig:ent_eng_sst}) in all datasets (e.g., QNLI). 

\begin{table}[]
\small
\centering
\begin{tabular}{l}
\hline
Text: \textit{when the film ended, i felt tired and drained and} \\ \textit{wanted to lie on my own deathbed.} Label: 1 \\
$\hat{E}_\theta(x)$: -9.37 Baseline: (.999, .001) $\rightarrow$ EBM: (.998, .002) \\
\hline
Text: \textit{sit through this one, you won't need a magic watch} \\ \textit{to stop time; your dvd player will do it for you.} Label: 1 \\
$\hat{E}_\theta(x)$: -7.57 Baseline: (.006, .994) $\rightarrow$ EBM: (.345, .655) \\
\hline
\end{tabular}
    \vspace{-0.2cm}
\caption{The change of the model's confidence (posterior distribution) for low and high-energy data samples of SST-2. The EBM variant shown is \textit{sharp-hidden}. We also provide QNLI examples in Appendix \ref{appsec:auxiliary_results}.}
\label{tab:energy_examples}
\end{table}

\section{Related Works}
Finally, we review applications of NCE or energy-based models in the NLP literature. Due to its self-normalizing property, NCE training has been used for faster inference \citep{nce12mnih,xiechen17ncernn,labeau-allauzen-2018-learning} of auto-regressive language models. It has also been used in an attempt to train a sentence-level  bi-directional LM \citep{ncebilm16tianxing}.

 More closely related to our work, \citet{Deng2020Residual} adopts NCE to train an EBM defined on top of a text encoder (the \textit{scalar} variant), and uses it to improve language generation. EBM has also been recently used in non-autoregressive machine translation \citep{tu-etal-2020-engine}.

\section{Conclusion}
In this work, we explore joint EBM training during the finetuning of pretrained text encoders with noise contrastive estimation. We find that joint EBM training can greatly improve the calibration of NLU models, with little or no loss on accuracy. 

\section*{Acknowledgements}
We sincerely thank Hao Tang and Will  Grathwohl for valuable discussions and suggestions.

\bibliography{eacl2021}

\begin{thebibliography}{21}
\expandafter\ifx\csname natexlab\endcsname\relax\def\natexlab#1{#1}\fi

\bibitem[{Chen et~al.(2015)Chen, Liu, Gales, and Woodland}]{xiechen17ncernn}
Xie Chen, Xunying Liu, Mark J.~F. Gales, and Philip~C. Woodland. 2015.
\newblock \href
  {http://dblp.uni-trier.de/db/conf/icassp/icassp2015.html#ChenLGW15a}
  {Recurrent neural network language model training with noise contrastive
  estimation for speech recognition.}
\newblock In \emph{ICASSP}, pages 5411--5415. IEEE.

\bibitem[{Deng et~al.(2020)Deng, Bakhtin, Ott, Szlam, and
  Ranzato}]{Deng2020Residual}
Yuntian Deng, Anton Bakhtin, Myle Ott, Arthur Szlam, and Marc'Aurelio Ranzato.
  2020.
\newblock \href {https://openreview.net/forum?id=B1l4SgHKDH} {Residual
  energy-based models for text generation}.
\newblock In \emph{International Conference on Learning Representations}.

\bibitem[{Devlin et~al.(2018)Devlin, Chang, Lee, and Toutanova}]{jacob18bert}
Jacob Devlin, Ming{-}Wei Chang, Kenton Lee, and Kristina Toutanova. 2018.
\newblock \href {http://arxiv.org/abs/1810.04805} {{BERT:} pre-training of deep
  bidirectional transformers for language understanding}.
\newblock \emph{CoRR}, abs/1810.04805.

\bibitem[{Fan et~al.(2018)Fan, Lewis, and Dauphin}]{fan2018-storyhierarchical}
Angela Fan, Mike Lewis, and Yann Dauphin. 2018.
\newblock \href {https://doi.org/10.18653/v1/P18-1082} {Hierarchical neural
  story generation}.
\newblock In \emph{Proceedings of the 56th Annual Meeting of the Association
  for Computational Linguistics (Volume 1: Long Papers)}, pages 889--898,
  Melbourne, Australia. Association for Computational Linguistics.

\bibitem[{Grathwohl et~al.(2019)Grathwohl, Wang, Jacobsen, Duvenaud, Norouzi,
  and Swersky}]{grathwohl2019classifier}
Will Grathwohl, Kuan-Chieh Wang, Jörn-Henrik Jacobsen, David Duvenaud,
  Mohammad Norouzi, and Kevin Swersky. 2019.
\newblock \href {http://arxiv.org/abs/1912.03263} {Your classifier is secretly
  an energy based model and you should treat it like one}.

\bibitem[{Guo et~al.(2017)Guo, Pleiss, Sun, and Weinberger}]{pmlr-v70-guo17a}
Chuan Guo, Geoff Pleiss, Yu~Sun, and Kilian~Q. Weinberger. 2017.
\newblock \href {http://proceedings.mlr.press/v70/guo17a.html} {On calibration
  of modern neural networks}.
\newblock volume~70 of \emph{Proceedings of Machine Learning Research}, pages
  1321--1330, International Convention Centre, Sydney, Australia. PMLR.

\bibitem[{Gutmann and Hyvärinen(2010)}]{pmlr-v9-gutmann10a}
Michael Gutmann and Aapo Hyvärinen. 2010.
\newblock \href {http://proceedings.mlr.press/v9/gutmann10a.html}
  {Noise-contrastive estimation: A new estimation principle for unnormalized
  statistical models}.
\newblock In \emph{Proceedings of the Thirteenth International Conference on
  Artificial Intelligence and Statistics}, volume~9 of \emph{Proceedings of
  Machine Learning Research}, pages 297--304, Chia Laguna Resort, Sardinia,
  Italy. PMLR.

\bibitem[{He et~al.(2015)He, Zhang, Ren, and Sun}]{He2015resnet}
Kaiming He, Xiangyu Zhang, Shaoqing Ren, and Jian Sun. 2015.
\newblock Deep residual learning for image recognition.
\newblock \emph{arXiv preprint arXiv:1512.03385}.

\bibitem[{He et~al.(2016)He, Zhang, Droppo, and Yu}]{ncebilm16tianxing}
Tianxing He, Yu~Zhang, Jasha Droppo, and Kai Yu. 2016.
\newblock \href {http://arxiv.org/abs/1602.06064} {On training bi-directional
  neural network language model with noise contrastive estimation}.
\newblock \emph{CoRR}, abs/1602.06064.

\bibitem[{Jung et~al.(2020)Jung, Kang, Cheng, Mentch, and
  Schaaf}]{jung-etal-2020-posterior}
Taehee Jung, Dongyeop Kang, Hua Cheng, Lucas Mentch, and Thomas Schaaf. 2020.
\newblock \href {https://doi.org/10.18653/v1/2020.acl-main.242} {Posterior
  calibrated training on sentence classification tasks}.
\newblock In \emph{Proceedings of the 58th Annual Meeting of the Association
  for Computational Linguistics}, pages 2723--2730, Online. Association for
  Computational Linguistics.

\bibitem[{Kingma and Ba(2014)}]{adam14kingma}
Diederik~P. Kingma and Jimmy Ba. 2014.
\newblock \href {http://arxiv.org/abs/1412.6980} {Adam: A method for stochastic
  optimization}.
\newblock Cite arxiv:1412.6980Comment: Published as a conference paper at the
  3rd International Conference for Learning Representations, San Diego, 2015.

\bibitem[{Kumar et~al.(2019)Kumar, Liang, and Ma}]{kuma19verifiedcal}
Ananya Kumar, Percy~S Liang, and Tengyu Ma. 2019.
\newblock \href
  {http://papers.nips.cc/paper/8635-verified-uncertainty-calibration.pdf}
  {Verified uncertainty calibration}.
\newblock In \emph{Advances in Neural Information Processing Systems 32}, pages
  3792--3803. Curran Associates, Inc.

\bibitem[{Labeau and Allauzen(2018)}]{labeau-allauzen-2018-learning}
Matthieu Labeau and Alexandre Allauzen. 2018.
\newblock \href {https://www.aclweb.org/anthology/C18-1261} {Learning with
  noise-contrastive estimation: Easing training by learning to scale}.
\newblock In \emph{Proceedings of the 27th International Conference on
  Computational Linguistics}, pages 3090--3101, Santa Fe, New Mexico, USA.
  Association for Computational Linguistics.

\bibitem[{LeCun et~al.(2006)LeCun, Chopra, Hadsell, Huang, and
  et~al.}]{LeCun06atutorial}
Yann LeCun, Sumit Chopra, Raia Hadsell, Fu~Jie Huang, and et~al. 2006.
\newblock A tutorial on energy-based learning.
\newblock In \emph{PREDICTING STRUCTURED DATA}. MIT Press.

\bibitem[{Liu et~al.(2019)Liu, Ott, Goyal, Du, Joshi, Chen, Levy, Lewis,
  Zettlemoyer, and Stoyanov}]{yinhan19roberta}
Yinhan Liu, Myle Ott, Naman Goyal, Jingfei Du, Mandar Joshi, Danqi Chen, Omer
  Levy, Mike Lewis, Luke Zettlemoyer, and Veselin Stoyanov. 2019.
\newblock \href {http://arxiv.org/abs/1907.11692} {Roberta: {A} robustly
  optimized {BERT} pretraining approach}.
\newblock \emph{CoRR}, abs/1907.11692.

\bibitem[{Ma and Collins(2018)}]{ma-collins-2018-noise}
Zhuang Ma and Michael Collins. 2018.
\newblock \href {https://doi.org/10.18653/v1/D18-1405} {Noise contrastive
  estimation and negative sampling for conditional models: Consistency and
  statistical efficiency}.
\newblock In \emph{Proceedings of the 2018 Conference on Empirical Methods in
  Natural Language Processing}, pages 3698--3707, Brussels, Belgium.
  Association for Computational Linguistics.

\bibitem[{Mnih and Teh(2012)}]{nce12mnih}
Andriy Mnih and Yee~Whye Teh. 2012.
\newblock A fast and simple algorithm for training neural probabilistic
  language models.
\newblock In \emph{Proceedings of the 29th International Coference on
  International Conference on Machine Learning}, ICML'12, page 419–426,
  Madison, WI, USA. Omnipress.

\bibitem[{Radford et~al.(2019)Radford, Wu, Child, Luan, Amodei, and
  Sutskever}]{radford2019language}
Alec Radford, Jeff Wu, Rewon Child, David Luan, Dario Amodei, and Ilya
  Sutskever. 2019.
\newblock Language models are unsupervised multitask learners.

\bibitem[{Tu et~al.(2020)Tu, Pang, Wiseman, and Gimpel}]{tu-etal-2020-engine}
Lifu Tu, Richard~Yuanzhe Pang, Sam Wiseman, and Kevin Gimpel. 2020.
\newblock \href {https://doi.org/10.18653/v1/2020.acl-main.251} {{ENGINE}:
  Energy-based inference networks for non-autoregressive machine translation}.
\newblock In \emph{Proceedings of the 58th Annual Meeting of the Association
  for Computational Linguistics}, pages 2819--2826, Online. Association for
  Computational Linguistics.

\bibitem[{Wang et~al.(2018)Wang, Singh, Michael, Hill, Levy, and
  Bowman}]{wang-etal-2018-glue}
Alex Wang, Amanpreet Singh, Julian Michael, Felix Hill, Omer Levy, and Samuel
  Bowman. 2018.
\newblock \href {https://doi.org/10.18653/v1/W18-5446} {{GLUE}: A multi-task
  benchmark and analysis platform for natural language understanding}.
\newblock In \emph{Proceedings of the 2018 {EMNLP} Workshop {B}lackbox{NLP}:
  Analyzing and Interpreting Neural Networks for {NLP}}, pages 353--355,
  Brussels, Belgium. Association for Computational Linguistics.

\bibitem[{Zadrozny and Elkan(2001)}]{Zadrozny01obtainingcalibrated}
Bianca Zadrozny and Charles Elkan. 2001.
\newblock Obtaining calibrated probability estimates from decision trees and
  naive bayesian classifiers.
\newblock In \emph{In Proceedings of the Eighteenth International Conference on
  Machine Learning}, pages 609--616. Morgan Kaufmann.

\end{thebibliography}
\bibliographystyle{acl_natbib}

\clearpage
\appendix
\section*{Appendices}
\section{Derivation of the \textit{hidden} Variant}
\label{appsec:hidden_derivation}
Remember from Section \ref{sec:notations}, the posterior distribution is obtained from a softmax operation on the logits, in other words:
\begin{equation}
\label{eq:cond_ygivenx}
    P_\theta(y|x) \propto \exp(f_\text{CLS}(enc(x))[y]).
\end{equation}

Without changing any parameters, one can re-use the logits to define an energy
based model of the joint distribution of data point x and labels y via:
\begin{equation}
\label{eq:jointxy}
    P_\theta(x, y) = \frac{\exp(f_\text{CLS}(enc(x))[y])}{Z(\theta)},
\end{equation}
where $Z(\theta)$ is the normalizing factor.
Note that Equation \ref{eq:jointxy} is \textbf{consistent} with Equation \ref{eq:cond_ygivenx} in the sense that Equation \ref{eq:cond_ygivenx} is a direct consequence of Equation \ref{eq:jointxy}.

Now by marginalizing out $y$, we get:
\begin{equation}
    P_\theta(x)= \frac{\sum^{|Y|}_{y=1}\exp(f_\text{CLS}(enc(x))[y])}{Z(\theta)},
\end{equation}
which is equivalent to 
\begin{equation}
    P_\theta(x)= \frac{\exp (- E_{\theta}(x))}{Z(\theta)},
\end{equation} where 
\begin{equation}
    E_{\theta}(x)=-\text{LogSumExp}_y(f_{\text{CLS}}(\text{enc}(x))[y]).
\end{equation}
For more intuition behind this derivation we refer readers to \citet{grathwohl2019classifier}.

\section{Details About the Noise Distribution}
\label{appsec:lmnoise}

We show some examples of generated noise samples and the masking in Table \ref{tab:lmnoise_sample_sst2}. Note that the masks could be applied to a consecutive span of words (Masking is applied to each token independently with probability $M$).

\begin{table}[h]
\small
\centering
\begin{tabular}{l}
\hline
\textbf{Input:} absolutely and completely \texttt{<M>} (\texttt{ridiculous}) \\
\textbf{Gen:} absolutely and completely hilarious \\
\hline
\textbf{Input:} \texttt{<M>} (\texttt{as a})  young \texttt{<M>} (\texttt{woman}) of great charm, \\ \texttt{<M>} (\texttt{generosity}) and diplomacy      \\
\textbf{Gen:} of a young man with a great charm, wit and \\ diplomacy \\
\hline
\end{tabular}
\caption{Example of generated noise samples on SST-2. The original words that are masked are also shown.}
\label{tab:lmnoise_sample_sst2}
\end{table}

Another possible way to get noise samples is that we can sample from BERT or Roberta with masked input. However, due to the nature of masked language modeling and the architecture of BERT / Roberta, the sampled tokens will be independent of each other, which could result in unnatural noise samples. That is why we choose to utilize an auto-regressive LM (e.g., GPT-2).

\section{Definition of ECE}
\label{appsec:ece_def}
Given an input sample $x$, for each label $y$, we say that the model predicts that $x$ belongs to label $y$ with confidence $P_\theta(y|x)$. Assuming the test-set contains $n$ samples, we will have $n \times |Y|$ predictions. 

ECE first partitions all predictions into $B$ equally-spaced bins by its confidence. Following \citet{jung-etal-2020-posterior, grathwohl2019classifier}, we set $B=20$, which means the width of each bin is 0.05. For example, the first bin contains all predictions that have confidence in the range of $[0, 0.05)$. Then for each bin ECE computes how the average of confidence is different from its actual accuracy: 
\begin{equation}
\small
\text{ECE} = \frac{1}{|Y|} \sum^{|Y|}_{y=1}\sum^B_{b=1}\frac{|B_{yb}|}{n}|\text{acc}(B_{yb})-\text{conf}(B_{yb})|,
\end{equation}
where $n$ is the number of samples in the test set, and $\text{acc}(B_{yb})$ is simply the ratio of samples ($x$) whose true label is indeed $y$ in $B_{yb}$.

\section{Auxiliary Results and Examples}
\label{appsec:auxiliary_results}

Examples of the model's confidence for low and high-energy data samples in QNLI are shown in Table \ref{tab:qnli_energy_sample}.

\begin{table*}[h]
\small
\centering
\begin{tabular}{l}
\hline
Text: \textit{Q: What city north of New York was settled by} \\ \textit{Huguenots? A: Huguenot immigrants did not disperse} \\ \textit{or settle in different parts of the country, but rather,} \\ \textit{formed three societies or congregations; one in the city of} \\ \textit{New York, another 21 miles north of New York} \\ \textit{in a town which they named New Rochelle, and} \\ 
\textit{a third further upstate in New Paltz.} Label: 1 \\
$\hat{E}_\theta(x)$: -8.48 Baseline: (.997, .003) $\rightarrow$ EBM: (.995, .005) \\
\hline
Text: \textit{Q: What is the source of oxygen production through} \\ \textit{electrocatalytic means? A: A similar method is the} \\ \textit{electrocatalytic  O2 evolution from oxides} \\ \textit{and oxoacids.} Label: 1 \\
$\hat{E}_\theta(x)$: 4.22 Baseline: (.252, .748) $\rightarrow$ EBM: (.472, .527) \\
\hline
\end{tabular}
\caption{The change of the model's confidence (posterior distribution) for low and high-energy data samples in the test set of QNLI. The EBM variant shown is \textit{sharp-hidden}.}
\label{tab:qnli_energy_sample}
\end{table*}

\begin{figure*}[h]
    \centering
    \includegraphics[width=0.6\linewidth]{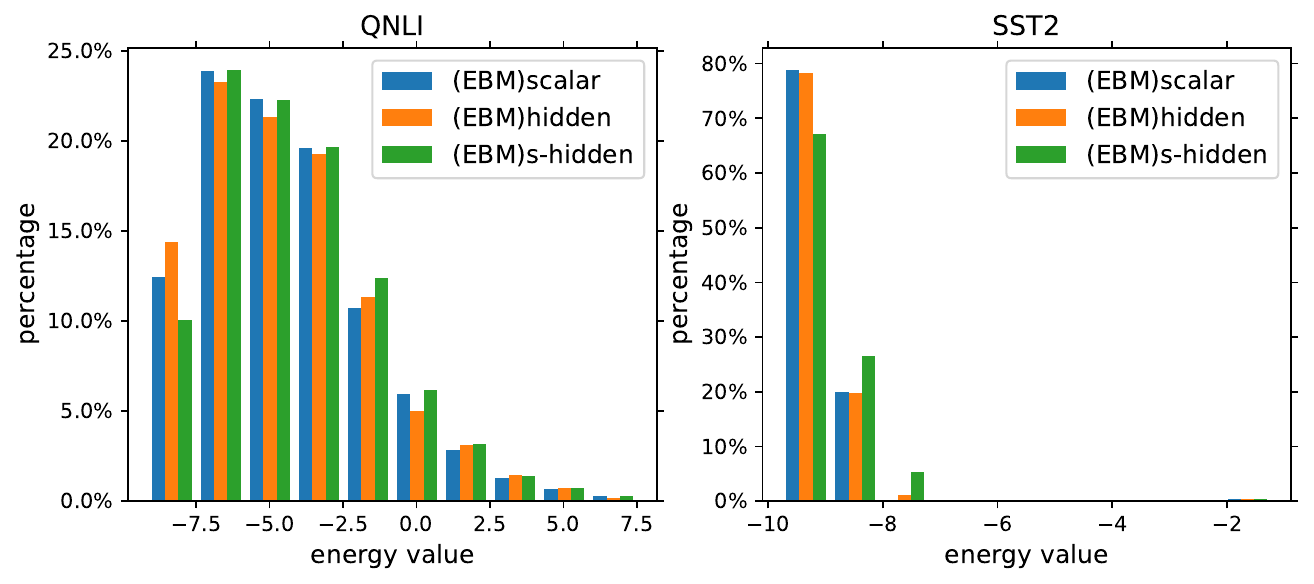}
    \caption{The histogram of energy values $\hat{E}_\theta(x)$ for samples in the test set of QNLI and SST-2.}
    \label{fig:hist_energy_qnli_sst2}
\end{figure*}

The histogram of energy values $\hat{E}_\theta(x)$ for samples in the test set of QNLI and SST-2 are shown in Figure \ref{fig:hist_energy_qnli_sst2}.

\begin{figure*}
    \centering
    \includegraphics[width=0.7\linewidth]{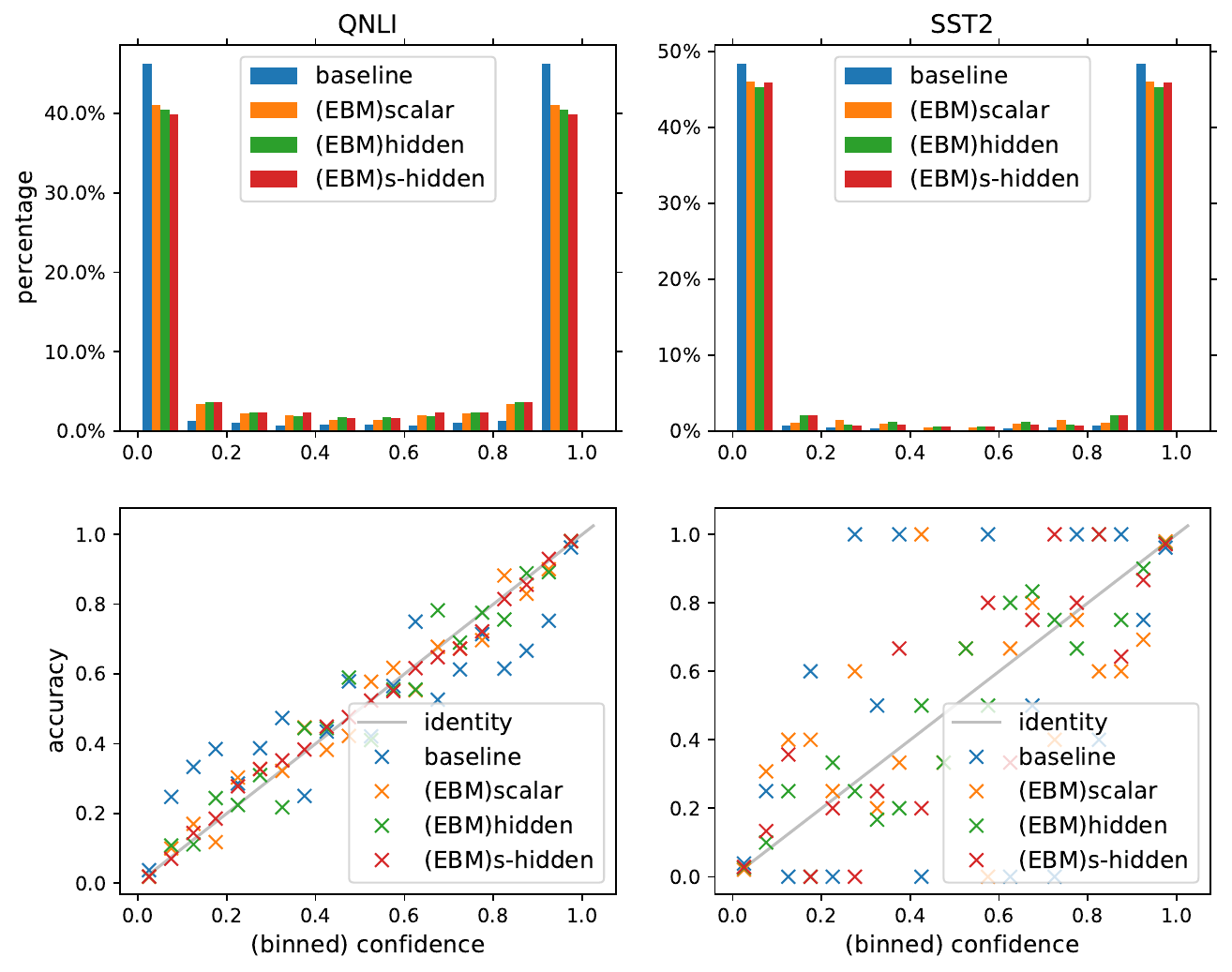}
        \vspace{-0.2cm}
    \caption{Visualization of calibration on QNLI and SST-2. Enlarged version.}
    \label{fig:large_cal_qnli_rte}
\end{figure*}

In Figure \ref{fig:large_cal_qnli_rte}, we provide an enlarged version of Figure \ref{fig:cal_qnli_rte}.

\end{document}